# An Investigation of Supervised Learning Methods for Authorship Attribution in Short Hinglish Texts using Char & Word N-grams.

ABHAY SHARMA, Microsoft Corporation, India
ANANYA NANDAN, Guru Gobind Singh Indraprastha University, India
REETIKA RALHAN, Guru Gobind Singh Indraprastha University, India

The writing style of a person can be affirmed as a unique identity indicator; the words used, and the structuring of the sentences are clear measures which can identify the author of a specific work. Stylometry and its subset — Authorship Attribution, have a long history beginning from the $19^{th}$ century, and we can still find their use in modern times. The emergence of the Internet has shifted the application of attribution studies towards non-standard texts that are comparatively shorter to and different from the long texts on which most research has been done. The aim of this paper focuses on the study of short online texts, retrieved from messaging application called WhatsApp and studying the distinctive features of a macaronic language (Hinglish), using supervised learning methods and then comparing the models. Various features such as word n-gram and character n-gram are compared via methods viz., Naïve Bayes Classifier, Support Vector Machine, Conditional Tree, and Random Forest, to find the best discriminator for such corpora. Our results showed that SVM attained a test accuracy of up to 95.079% while similarly, Naïve Bayes attained an accuracy of up to 94.455% for the dataset. Conditional Tree & Random Forest failed to perform as well as expected. We also found that word unigram and character 3-grams features were more likely to distinguish authors accurately than other features.

CCS Concepts: • **Computing methodologies → Information extraction**; **Language resources**; *Supervised learning*; *Supervised learning by classification*; Support vector machines;

Additional Key Words and Phrases: authorship attribution, macaronic languages, Hinglish, stylometry, machine learning, social media, WhatsApp

## 1 INTRODUCTION

With accelerated evolution of the internet, the online textual material present before us has also increased with time. The anonymous nature of the abundant data which is easily available gives rise to illicit possibilities. Thus, determining the authors of some unknown texts for verification purposes is a modern-day need. Such an attempt to identify the author of a given text using Natural Language

Authors' addresses: Abhay Sharma, Microsoft Corporation, ISB Road, Gachibowli, Hyderabad, 500032, India, abhay@protonmail.com; Ananya Nandan, Guru Gobind Singh Indraprastha University, Maharaja Surajmal Institute of Technology, C-4, Janakpuri, Delhi, New Delhi, 110058, India; Reetika Ralhan, Guru Gobind Singh Indraprastha University, Maharaja Surajmal Institute of Technology, C-4, Janakpuri, Delhi, New Delhi, 110058, India.

Processing (NLP) is known as Authorship Attribution. With the help of basic textual features like word or sentence length, we can determine the author of a given text. But the extraction of only such textual features does not provide us with accurate results since every author's linguistic style greatly varies. Ergo, to define the style of an author, we need to study and quantify the linguistic style of an author. This quantitative method used is termed as stylometry. Stylometry emphasizes on the study of the writing style of a certain author, which highlights certain features that are independent of the author's will and cannot be manipulated.

While Authorship Attribution in English has gained attention like the works of Efstathios Stamatatos, Nikos Fakotakis, and George Kokkinakis on automatic text categorization in terms of genre and author [25] or authorship attribution by Patric Juola [11–13]. One such work is by Roy Schwartz, Oren Tsur, Ari Rappoport and Moshe Koppel on Authorship Attribution of Micro Messages [22], mainly focusing on tweets. But the focus of our research is *macaronic languages*. These languages are a mixture of languages. For example, the macaronic language formed by the switching from English to Hindi (or other Indian dialects) is termed as Hinglish, i.e., it contains parts of both the languages. Although researches have been conducted in the past on macaronic languages by Rahel Oppliger on Automatic Authorship Attribution based on character n-gram as features of Naïve Bayes Classifier, in Swiss-German [19], the work done under the umbrella of such languages is quite low. The non-standard features of these languages, like subject-specific spellings and inter-mixed grammar standards, give rise to a large number of distinctive attributes for various authors.

The main focus of our paper is concentrated on the analysis of one of such language — Hinglish. Since the corpus to be used for our research is difficult to gather due to low availability (as Hinglish is region specific and nonstandard), we decided to use texts from a popular messaging platform — WhatsApp.

The next section describes the corpora that have been used for the process of authorship attribution while section 3 describes the methodology. The main focus lies on the study of word n-gram and character n-gram Naïve Bayes Classifier, Support Vector Machine, Conditional Tree, and Random Forest. Our aim is to understand and compare the methods. Section 4 depicts the results extracted from the methods used in Section 3. With the results in hand, Section 5 holds the discussions based on the results obtained that further leads to the conclusions of the results, which are described in Section 6.

## 2 CORPORA

The data used in this study was obtained from an Instant Messaging Application — WhatsApp; and includes personal as well as group texts from four different authors. The dataset consisted of 76,000 words (approximately uniformly distributed). All the participants shared a similar dialect. The conversation they conducted was a combination of two languages — Hindi and English, switching between both as per their needs. One of the unique aspects of South Asian language users such as those writing in Hindi, Punjabi, Bengali etc. is the use of Latin script to depict the native language rather than the native script. So, for example, rather than using Devanagari script for Hindi, the user/authors use English alphabets which resonate to a similar sound and tone of a given word.

## 3 METHODOLOGY

### 3.1 Word n-gram

A word n-gram can be defined as a continuous sequence of n words. Word n-grams have been proposed as textual features by Peng, et al. [20] ; Sanderson & Guenther [21] ; Coyotl-Morales, Villaseñor-Pineda, Montes-y-Gómez, & Rosso [5]. However, the accuracy attained by word n-grams shows limitations in comparison to individual word features, or word unigrams (Sanderson &

Guenther [21] ; Coyotl-Morales, et al. [5]). The dimensionality of the problem increases if n-grams are used since the number of comparisons to be taken into consideration increases. Also, the output representation by this method is extremely sparse, since the majority of the combinations do not even occur in short texts. Moreover, it is very much possible to apprehend content-specific information instead of stylistic information [7].

To avoid such limitations of word n-grams, just word unigrams & bigrams are extracted from the texts. In unigram condition, the words are the primary features, and hence the words from the texts have to be isolated. A list of single words occurring in the texts is generated. For each text of each author, the occurrence of the feature, or word, is counted, i.e., the frequency of occurrence is calculated. This frequency is then normalized according to the text length. Normalization is necessary at this step, because the frequency of a given feature may be higher compared to other texts. If a text consists of a higher number of words, then the chance of occurrence of a particular feature in the said text expands. In bigram condition, the same methodology is used, except the number of words is two.

### 3.2 Character n-gram

In this family of measures, the text is regarded as a sequence of characters. This information regarding characters is easily available for any natural language and its corpus and is exceedingly beneficial to quantify the style of an author [8]. A more elaborate, yet computationally simplistic procedure is to retrieve frequency of n-grams on the character level. This can capture the nuances of the author's style. In addition, this method of representation is tolerant to noise. In certain cases, in which the texts under study are noisy — containing grammatical errors, or strange use of punctuation, the representation of character n-grams is not affected highly. Also, for oriental languages in which it is difficult to carry out the tokenization process, character n-grams are usually preferred [17]. The process involving the retrieval of most frequent n-grams is certainly language-independent and requires no special tools. However, the dimensionality of this depiction is very much increased to the word-based approach [23, 24], because character n-grams encapsulate redundant information as well, as well as many characters n-grams are required to represent an individual word. The applications of character n-gram approach have proven to be quite advantageous and successful in the arena of authorship attribution. Kjell [14] initially used character bigrams and trigrams to recognize and differentiate the Federalist Papers. Similarly, Forsyth and Holmes [6] proved that bigrams and character n-grams of variable length are a much better technique in comparison to lexical features in text classification tasks. In our research, since the focus lies on macaronic languages, more specifically on Hinglish, the words generally used by the authors constituted of a sequence of four or five letters. It was presumed that character 3-grams, 4-grams and 5-grams will be able to provide better and accurate results than other values of n.

### 3.3 Naïve Bayes Classifier

The Naïve Bayes classifier has demonstrated itself to be exceptional for most of the text processing experiments including text classification. The first publication, dealing with Bayesian methods as applied to large-scale data analysis was probably carried out by Mosteller et al. [18] to classify and provide statistical evidence on the most probable author of The Federalist papers. More recently, Hoorn et al. [9] identified the intended poet behind the definite prose using letter sequences through neural networks. Peng et al. [20] proposed to augment the Naïve Bayes models with statistical n-gram language models, thereby removing the shortcomings of the standard Naïve Bayes Text Classifier.

Given a set of features F= {f1, f2, f3. ....and so on}, Naïve Bayes can be utilized to determine the probability of a document $d$ belonging to the class $c_i$ , as follows [24] :

$$P(c_i|d) = \frac{P(d|c_i)P(c_i)}{P(d)}$$
$$P(c_i|d) = P(c_i)\ \pi_{f=i\ to\ |F|}\ P(f_i|c_i)$$

### 3.4 Support Vector Machine

Based on Statistical Learning Theory and Structural Risk Mitigation [27], Support Vector Machines were first propounded by Vapnik for classification and forecasting purposes. They have been extensively used for studies involving text classification, pattern, speech and image recognition, face detection, etc. Thus, use of SVM in data mining applications makes it an incumbent tool for development of products in varied fields.

R Burbidge et al. [4] demonstrated and compared various machine learning methods already being used in structure-activity relationship analysis, with SVM, and observed that SVM outperformed all these techniques.

Giorgio Valentini [26] proposed to classify types of lymphoma and analyze the role of coordinately expressed genes (n-grams) in carcinogenic processes, using SVM. To reduce the training and testing error for better performances, Chin-Teng Lin et al. [16] have proposed an SVM based Fuzzy Neural Network, to develop an algorithm so that the clustering principle is able to determine the fuzzy rules and membership functions, automatically.

The principle idea for SVM is constructing the optimal hyperplane, used for classifying the linear separable patterns. In simpler words, a hyperplane is a plane chosen from a set of similar hyper planes that maximizes the margin of hyperplanes, so as to correctly classify the patterns.

Hyperplane can be represented by the following equation:

$$aX + bY = C$$

### 3.5 Conditional Tree

Conditional Tree Learning uses a Conditional Tree as a predictive model which maps observations about an item (represented in the branches) to conclusions about the item's target value (represented in the leaves). It is one of the predictive modeling approaches used in statistics, data mining and machine learning. Tree models where the target variable can take a finite set of values are called **classification trees**; in these tree structures, leaves represent class labels and branches represent conjunctions of features that lead to those class labels. Decision trees where the target variable can take continuous values (typically real numbers) are called **regression trees**. In decision analysis, a conditional tree can be used to represent decisions and decision-making visually and explicitly. In data mining, a conditional tree describes data (but the resulting classification tree can be an input for decision making). [1]

### 3.6 Random Forest

Random Forest generates a group of different decision trees [15]. It is basically a classifier which consists of a set of tree-structured classifiers, $\{h(x, \Theta_k), where\ k = 1,2..... \}$, where $\{ \Theta_k \}$ can be seen as an independent identically distributed random vector, and each tree of the set put forward of a vote for the most definite class at a given input $x$.

To achieve diversity among the set of decision trees, Breimann [2, 3] experimented through the following steps: The number of records $N$ given in the training set are samples randomly, which in turn is used as the training set for the growing tree. If there are $M$ input variables, a number $m << M$ is chosen so that at each node of the tree, m variables are chosen randomly out of $M$ and the best split on these m attributes is used to split the node. The value of m is kept consistent while the forest grows. Each tree is developed to the largest extent possible. Thus, multiple trees are induced in the forest; the number of trees is pre-decided by a new parameter *Ntree*. Once the forest is constructed, it is run across all the trees in the forest. Each tree gives classification for the new instance which is recorded as a vote (by each tree). The votes from all trees are amalgamated and the class for which maximum votes are enumerated is declared as classification of the new instance. This process is referred to as Forest RI [3].

### 3.7 Putting it all together

In our research, the short online texts taken from four different authors make up the corpora. These online texts are cleaned and converted into the desired encoding format. N-gram features such as Word n-gram (Unigram and Bigram) and Character n-gram (Character 3-gram, 4-gram, and 5-gram) are then extracted from the corpus, so as to compare across classifiers, viz. Naïve Bayes, Support Vector Machine, Conditional Tree and Random Forest. These supervised learning algorithms have been previously used in Authorship Identification experiments for standard languages. Our aim was to understand and compare these models for macaronic languages.

Our experiment begins by, firstly, splitting the corpus into training and testing data. 30% of the data makes up the training data, and the remaining 70% is split into half, thereby building two testing data. After splitting the data, the above-mentioned features are extracted from the corpus. However, all the features extracted may not be able to train and test the classifiers efficiently, hence proper feature selection is carried out using weightage methods, i.e., Term Frequency (TF), Term Frequency-Inverse Document Frequency (TF-IDF), and Binary Weight. By doing this, we are also able to compare these methods. The features finally are extracted in the form of a document matrix, and these are deployed to train the classifiers to build respective training models. Lastly, the two-testing data are tested on these models, for validating the performance of the resultant classification models. Evaluation techniques are assessed to estimate the future performance by measures such as accuracy, sensitivity, and precision, to maximize the empirical results.

The results of the two testing data are described in a tabular form in Section 4.

## 4 RESULTS

### 4.1 Naïve Bayes Classifier

Table 1. Result of Naïve Bayes Classification

| Weight | Token | Test One | | | Test Two | | |
|---|---|---|---|---|---|---|---|
| | | *Accuracy* | *TPR* | *Precision* | *Accuracy* | *TPR* | *Precision* |
| TF | Unigram (Word) | 81.890 | 77.268 | 79.788 | 81.225 | 75.741 | 80.907 |
| | Bigram (Word) | 68.701 | 65.115 | 54.211 | 67.194 | 67.083 | 53.096 |
| | Char 3 Gram | 77.362 | 76.439 | 74.302 | 75.692 | 71.858 | 73.807 |
| | Char 4 Gram | 79.528 | 77.478 | 75.924 | 78.459 | 76.055 | 78.769 |
| | Char 5 Gram | 79.331 | 79.308 | 77.565 | 80.632 | 78.703 | 81.981 |
| TF-IDF | Unigram (Word) | 79.331 | 70.521 | 80.294 | 79.644 | 71.997 | 82.683 |
| | Bigram (Word) | 8.661 | 24.441 | 26.894 | 7.510 | 21.700 | 25.758 |
| | Char 3 Gram | 87.402 | 81.469 | 82.875 | 90.316 | 85.995 | 86.456 |
| | Char 4 Gram | 80.315 | 70.918 | 82.030 | 80.632 | 72.474 | 79.424 |
| | Char 5 Gram | 55.709 | 53.466 | 62.757 | 54.150 | 52.589 | 58.246 |

*Continued on next page*

*Table 1 continued*

| | | | | | | | |
|---|---|---|---|---|---|---|---|
| Weight Bin | Unigram (Word) | 94.276 | 93.856 | 91.817 | 94.455 | 93.827 | 93.588 |
| | Bigram (Word) | 69.685 | 64.719 | 56.738 | 68.577 | 67.460 | 54.405 |
| | Char 3 Gram | 93.194 | 92.430 | 91.372 | 93.259 | 93.029 | 92.068 |
| | Char 4 Gram | 91.929 | 95.155 | 90.444 | 91.304 | 92.841 | 89.528 |
| | Char 5 Gram | 92.701 | 92.125 | 90.405 | 92.061 | 91.248 | 90.974 |

## 4.2 Support Vector Machine

Table 2. Result of SVM Classification

| Weight | Token | Test One | | | Test Two | | |
|---|---|---|---|---|---|---|---|
| | | *Accuracy* | *TPR* | *Precision* | *Accuracy* | *TPR* | *Precision* |
| TF | Unigram (Word) | 91.535 | 94.444 | 82.006 | 94.269 | 96.623 | 88.787 |
| | Bigram (Word) | 69.291 | 74.838 | 46.673 | 68.972 | 74.466 | 45.090 |
| | Char 3 Gram | 91.929 | 92.494 | 81.825 | 92.095 | 94.935 | 83.038 |
| | Char 4 Gram | 90.551 | 93.208 | 77.028 | 91.107 | 95.016 | 80.089 |
| | Char 5 Gram | 91.929 | 95.578 | 80.881 | 90.514 | 93.798 | 79.471 |
| TF-IDF | Unigram (Word) | 78.937 | 93.245 | 55.416 | 76.087 | 92.604 | 54.875 |
| | Bigram (Word) | 60.630 | 28.048 | 32.508 | 59.684 | 27.690 | 32.986 |
| | Char 3 Gram | 93.307 | 96.658 | 84.544 | 92.885 | 96.401 | 84.838 |
| | Char 4 Gram | 83.071 | 94.267 | 62.067 | 83.992 | 94.512 | 68.419 |
| | Char 5 Gram | 62.992 | 40.147 | 30.871 | 61.660 | 39.938 | 29.545 |
| Weight Bin | Unigram (Word) | 94.094 | 96.865 | 86.091 | 94.862 | 97.061 | 89.404 |
| | Bigram (Word) | 70.866 | 78.645 | 47.655 | 69.565 | 80.337 | 45.589 |
| | Char 3 Gram | 95.079 | 97.781 | 87.172 | 94.676 | 97.248 | 85.477 |
| | Char 4 Gram | 94.291 | 97.490 | 85.003 | 92.885 | 96.730 | 83.948 |
| | Char 5 Gram | 93.504 | 96.555 | 84.130 | 93.083 | 96.373 | 85.471 |

## 4.3 Conditional Tree

Table 3. Result of Conditional Tree Classification

| Weight | Token | Test One | | | Test Two | | |
|---|---|---|---|---|---|---|---|
| | | *Accuracy* | *TPR* | *Precision* | *Accuracy* | *TPR* | *Precision* |
| TF | Unigram (Word) | 74.604 | 50.315 | 61.320 | 73.518 | 51.153 | 63.614 |
| | Bigram (Word) | 67.126 | 70.803 | 44.866 | 66.996 | 76.619 | 46.251 |
| | Char 3 Gram | 78.346 | 56.929 | 57.710 | 80.435 | 59.425 | 69.451 |
| | Char 4 Gram | 79.528 | 57.902 | 64.986 | 79.842 | 58.764 | 68.542 |
| | Char 5 Gram | 73.819 | 50.825 | 61.180 | 73.320 | 51.065 | 64.278 |
| TF-IDF | Unigram (Word) | 75.197 | 62.684 | 58.386 | 75.099 | 72.850 | 59.896 |
| | Bigram (Word) | 65.157 | 70.778 | 42.869 | 64.625 | 76.966 | 43.875 |
| | Char 3 Gram | 76.378 | 57.228 | 64.283 | 77.470 | 58.701 | 67.057 |

*Continued on next page*

*Table 3 continued*

| | | | | | | | |
|---|---|---|---|---|---|---|---|
| | Char 4 Gram | 75.390 | 66.808 | 53.026 | 74.111 | 46.756 | 48.721 |
| | Char 5 Gram | 75.591 | 56.959 | 59.364 | 74.704 | 58.300 | 60.108 |
| Weight Bin | Unigram (Word) | 75.000 | 52.102 | 62.997 | 75.494 | 54.891 | 66.292 |
| | Bigram (Word) | 67.323 | 70.868 | 45.055 | 66.798 | 76.558 | 46.061 |
| | Char 3 Gram | 78.740 | 58.311 | 65.012 | 79.644 | 59.094 | 67.056 |
| | Char 4 Gram | 78.346 | 57.745 | 57.877 | 79.051 | 58.970 | 63.655 |
| | Char 5 Gram | 75.197 | 54.461 | 62.434 | 75.296 | 56.773 | 64.362 |

## 4.4 Random Forest

Table 4. Result of Random Forest Classification

| Weight | Token | Test One | | | Test Two | | |
|---|---|---|---|---|---|---|---|
| | | *Accuracy* | *TPR* | *Precision* | *Accuracy* | *TPR* | *Precision* |
| TF | Unigram (Word) | 59.449 | 39.596 | 27.462 | 59.486 | 64.604 | 30.192 |
| | Bigram (Word) | 60.433 | 39.745 | 28.409 | 58.103 | 39.400 | 26.136 |
| | Char 3 Gram | 58.268 | 64.421 | 29.055 | 58.696 | 64.487 | 31.072 |
| | Char 4 Gram | 57.874 | 39.364 | 25.947 | 57.905 | 64.371 | 27.039 |
| | Char 5 Gram | 56.890 | 14.222 | 25.000 | 56.920 | 14.229 | 25.000 |
| TF-IDF | Unigram (Word) | 63.780 | 65.275 | 32.175 | 63.439 | 65.222 | 35.071 |
| | Bigram (Word) | 60.039 | 39.685 | 28.030 | 57.708 | 39.343 | 25.758 |
| | Char 3 Gram | 66.535 | 65.741 | 35.372 | 67.984 | 66.000 | 39.427 |
| | Char 4 Gram | 58.465 | 39.450 | 26.515 | 58.300 | 64.429 | 27.418 |
| | Char 5 Gram | 59.055 | 39.537 | 27.083 | 58.103 | 64.400 | 26.682 |
| Weight Bin | Unigram (Word) | 61.417 | 64.897 | 29.902 | 60.277 | 64.724 | 29.311 |
| | Bigram (Word) | 60.433 | 39.745 | 28.409 | 58.300 | 39.429 | 26.326 |
| | Char 3 Gram | 57.283 | 64.279 | 25.925 | 58.103 | 39.400 | 29.412 |
| | Char 4 Gram | 56.335 | 39.279 | 25.379 | 57.312 | 39.286 | 26.471 |
| | Char 5 Gram | 56.248 | 39.190 | 25.370 | 57.410 | 64.314 | 26.114 |

# 5 DISCUSSION

The research in the area of authorship attribution has been largely done on standard languages, due to easy availability of corpora and the extensive features that are accessible to study authorship for such languages. The study of macaronic languages is a difficult one since extensive datasets are often unavailable for research and moreover, there are no proper tools to process the features. With our project, we wished to understand the efficiency of modern attribution methods on one of such non-standard language and the intricacies of such languages.

Our investigation gave important understanding on the feasibility of utilizing standard strategies on macaronic dialects, and we found that such dialects are morphologically rich and have certain features like atypical spellings, gender explicit pronouns and user determined syntax, which differentiate them from standard dialects.

A distinctive feature of macaronic language seems to be their idiolectic spellings. Our Hinglish texts were also plagued by such spelling usage. To illustrate how idiolectic spellings are reflected, we look at distinctive word unigrams of authors: (see Figure 1).

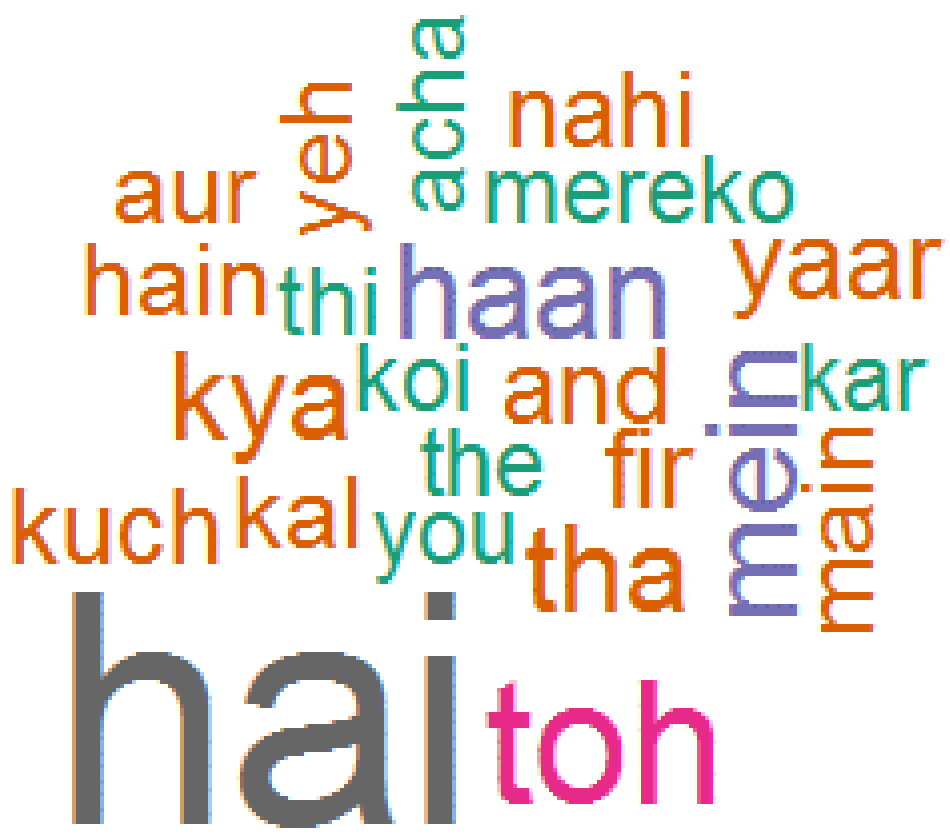

Fig. 1. Word cloud of top unigram words

A closer investigation of this word cloud reveals that words such as “mein” and “main” which can be used in interchangeably (and are just different Devanagari transliteration) can be highly indicative of a particular author since they represent a unique stylometric feature of an author. Words such as “hai” and “hain” can provide an important distinction because they might represent accent of an author that is embodied in text unknowingly.

Using Word Bigrams, we can observe more distinctive features: (see Figure 2).

The spellings of words -"raha" and "rha", “hai” and “h” which are different Devanagari transliteration act as an important differentiator. We can also see the distinctive grammar patterns, which are found in Hinglish.

Overall, consistent idiolectic spelling and grammar choices such as the aforementioned are shown to be indicative of authorship, especially if they are characteristic of only the specific author. In this way, these characteristic orthographical freedoms in Hinglish can be exploited as an effective feature for automatic authorship attribution. N-grams investigated in the study were successful in capturing the orthographical idiosyncrasies. As previously identified by Houvardas and Stamatatos [10], the major advantage of using n-gram features was their independence from strict grammar rules. Hence, they will play a vital role in authorship attribution in Macaronic Languages (or, in general, non-standard languages), since these languages do not follow, or more often, do not have any grammatical rules. The authors under our experiment have conducted their WhatsApp conversations in the same language — Hinglish, still they depict considerable differences in their grammatical style. Since such non-standard languages do not have strict grammar rules and spellings, the writing style of a particular author may differ from other, even if the meaning is clearly understood by the recipient of the text.

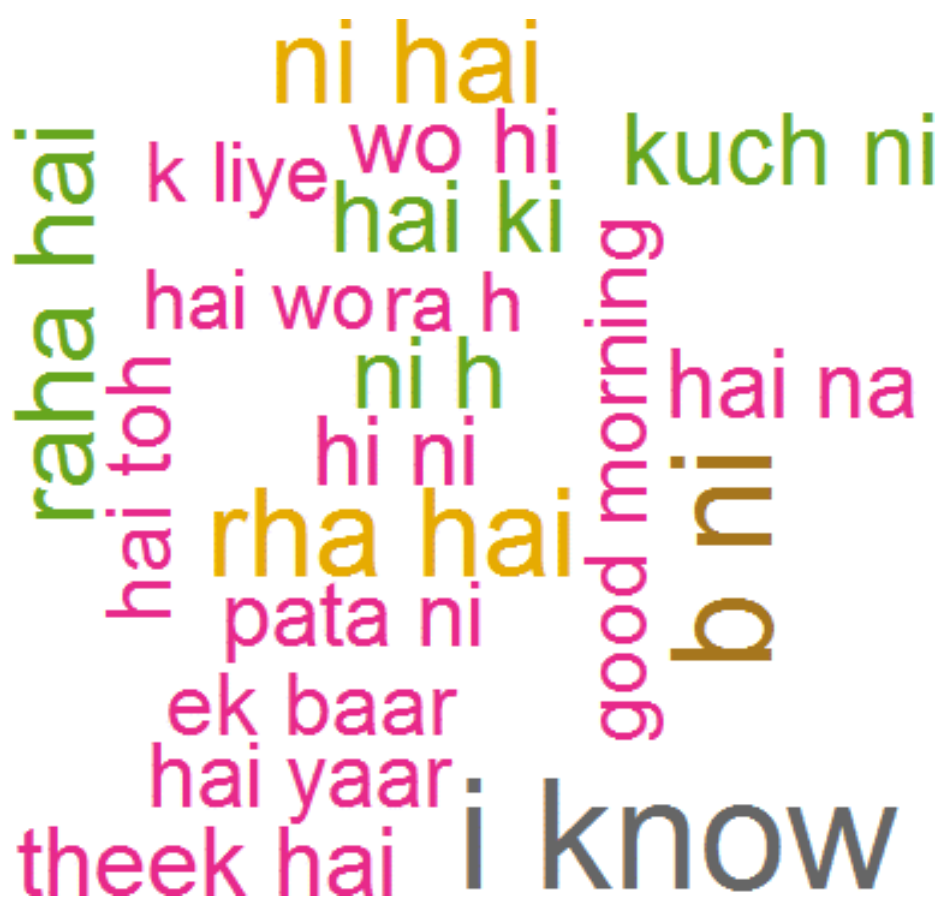

Fig. 2. Word cloud of top bigram words

It is clearly observed from the results that we have computed - Support Vector Machine, and Naïve Bayes Classifier provide much better Accuracy, TPR, and Precision, across all the features examined, in comparison to Conditional Tree and Random Forest. SVM provides accuracy of 94.862% in the case of Word unigram, and an accuracy of 95.079% in the case of Character 3-gram. Similarly, Naïve Bayes Classifier computed the maximum accuracy of 94.455% and 93.259%, in the case of Word unigram and Character 3-gram, respectively. Ergo, overall, SVM is considered to be the better classification method for Authorship Identification, among other algorithms under the study.

Moreover, from the above results, two observations can be duly noted. First, the better features for authorship identification for Hinglish are, unquestionably — Word unigram and Character 3-gram. Performance generally degraded when higher n-grams were used. Word bigrams led to surprisingly large performance reductions with accuracy decreasing up to 60%. This may be attributed to the sparsity of data set used and how the texts are too short for regular re-occurrence of bigrams. Second, Weight Bin generally presented the most optimal results out of all the three weightage techniques we assessed for our research. Also, Term Frequency was better than TF-IDF in algorithms like SVM and Naïve Bayes while TF-IDF worked better with algorithms like Decision Tree and Random Forest.

## 6 CONCLUSION

The aim of this research was to determine which types of information make it possible to identify the author of a short digital text and which supervised methods are better suited for a macaronic language such as Hinglish. We constructed several classifiers and compared them for achieving our goal.

The results indicate that SVM and Naïve Bayes work better than Conditional Tree and Random Forest in classifying texts. Overall, SVM performed better than all the algorithms used. Binary Weights were found to be the best representation for achieving high accuracy of classification, followed by Term Frequency and TF-IDF. We also found that Character 3-grams and Word unigram were better features for identifying and distinguishing authors.

In the end, our analysis gave valuable insight on the viability of using standard methods on macaronic languages, and we found that such languages are morphologically rich and have certain features such as atypical spellings, gender-specific pronouns, and user-specific grammar, which distinguish them from standard languages.

The outcome of this research leads us to conclude that lexical and character n-gram based authorship attribution using supervised methods in macaronic languages such as Hinglish is promising, even for datasets like instant messages.